\documentclass{OAGM}
\OAGMarXiv{1505.01065}

\usepackage{graphics}

\title{VeinPLUS: A Transillumination and Reflection-based Hand Vein Database}

\author{Alexander Gruschina\\
  Department of Computer Sciences, University of Salzburg, Austria}

\begin{document}
\maketitle

\begin{abstract}
  This paper gives a short summary of work related to the creation of a department-hosted hand vein database. After the introducing section, special properties of the hand vein acquisition are explained, followed by a comparison table, which shows key differences to existing well-known hand vein databases. At the end, the ROI extraction process is described and sample images and ROIs are presented.
\end{abstract}

\section{Introduction}
\begin{figure}[ht]
 \centering\scalebox{.3}{\includegraphics
 {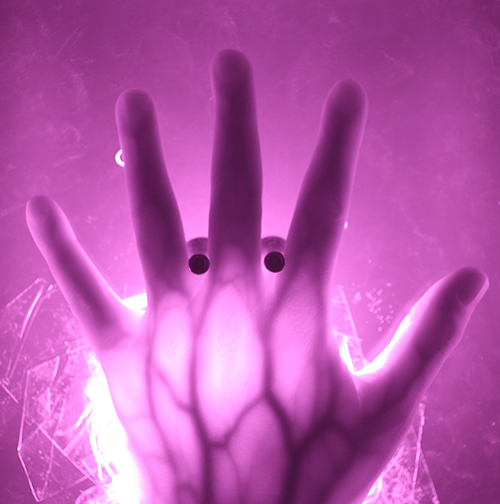}}\\
 \caption{Veins in the human hand}
 \label{fig:veins}
 \end{figure}

In the past decade, a new feature gained attention in the area of Biometrics. In general, veins are blood vessels in the human body needed for  blood transport between the heart and limbs. These veins have a very special structure (see Figure \ref{fig:veins}) and are said to be unique for every human being.  
Hand veins are so called \textit{intrinsic biometrics}, i.e. they are inside the human body. To show the advantages of intrinsic over extrinsic biometrics, let's compare hand veins with classical finger print.
Finger print information is pretty easy to forge. There are many different surfaces, where a person can leave it's finger print, e.g. a glass front of a touchscreen, which is part of nearly every mobile phone available today. As 90's crime TV shows have dramatically shown, extraction of this finger information can be easily achieved. This information can be further used for authentication. As opposed to finger print, hand vein information can not be left behind, since the vein pattern is not at the skin's surface. \newline
Another important difference is, that finger print can be damaged or malformed due to injuries. This problem is frequently faced by climbers or other people, who often have hand injuries. Since the veins are located inside the hand, injuries do not affect the crucial pattern. At long last, the crucial biometrical information is kept unrevealed. Actually, our skin is the privacy shield for our veins. 

The human vein pattern is captured by using near infrared (NIR) illumination. This part of the electromagnetic spectrum 
ranges from about 750 nm to 1400 nm. Near infrared light is located beyond the visible light range, which means photographs taken in NIR are pseudocolor images.

 
As mentioned before, the veins are blood vessels located in our body. The arteries carry blood from the heart to the tissues, containing \textit{oxidized hemoglobin}. When reaching the tissues, the oxygen is consumed and now the veins transport blood back to the heart. The \textit{deoxygenated} hemoglobin has a special property, it absorbs long wave light, i.e. near infrared illumination. So, if we irradiate the hand with NIR light, and capture an image within this spectrum, we can identify dark regions. These regions are the veins we searched for, since here the light was absorbed, instead of reaching the capturing device. Figure \ref{fig:veins} shows these regions.

\section{Technical Content}
There are some well-established databases in the area of hand veins. Nevertheless, one of the major tasks in my bachelor project, was to create a department-hosted hand vein database. There are some reasons, why generating our own database was chosen. The most important aspect is the investigation of two illumination variants. Transmitted light and reflected light lead to very different image results and there is no hand vein database available, which focuses on this topic. Of course, having the full control over the database is also important. Since we operate the whole creation process, we are familiar with all kind of information needed for evaluating the data we acquired. This does not only make us safe in the knowledge, that the data was handled properly, but also accelerates data processing, since all information (regarding e.g. conditions while acquisition process) is available immediately. Of course, it is also necessary to have a large vein database at one's disposal, as there will be many different areas, this database will be used for (e.g. research, analysis, testing, etc). 

\subsection{Hand vein scanner}
\label{section:scanner}
\begin{figure}[ht]
\centering
\begin{minipage}{.5\textwidth}
  \centering
  \scalebox{.25}{\includegraphics{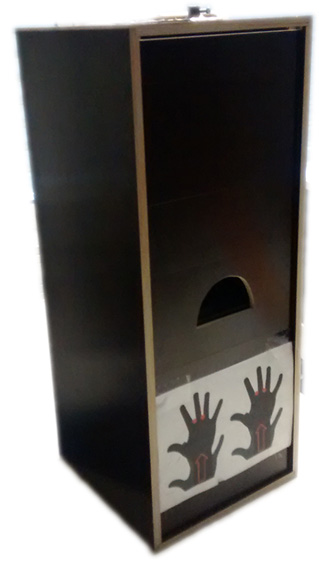}}
  \caption{Our hand vein scanner}
  \label{fig:scanner_front}
\end{minipage}%
\begin{minipage}{.5\textwidth}
  \centering
  \scalebox{.5}{\includegraphics{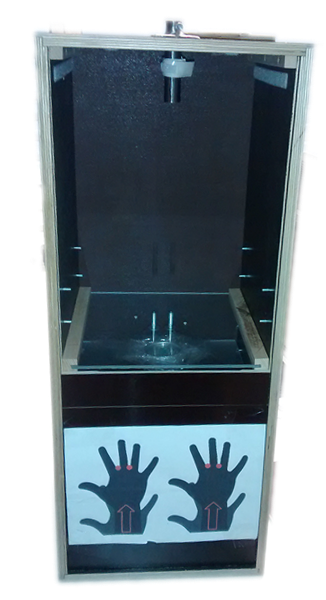}}
  \caption{Front panels are removable}
  \label{fig:scanner_opened}
\end{minipage}
\end{figure}

Probably the most crucial part while creating the database is the acquisition setup itself. Since one wants to have the same conditions for all captured images, we decided to use additional hardware, which guarantees us this requirement. Our \textit{hand vein scanner}, as we named it, is shown in Figure \ref{fig:scanner_front}. 

The scanner is a black colored wooden box, which was designed to enable an acquisition process, without having any disturbing signals while taking images of subjects' hands. The front of the box consists of removable panels (see Figure \ref{fig:scanner_opened}), which were thought for enabling different hand position heights. One of those panels has a semicircle-shaped aperture, where subjects position their hand in the device properly. During acquisition process, the hand rests on an acrylic glass plate, which is located behind the special shaped panel. As mentioned before, it is important that we have the same conditions for all subjects. One of these conditions is the hand's position. To handle variances in this area, two metallic pegs are screwed on the plate. These two pegs are not only advantageous to us, but also a help for the subject themselves. Since the scanner is indeed a \textit{black box}, an orientation help is a welcome feature.

On top of the scanner, another aperture can be identified. This one is intended for insertion of the imaging device. This device is mounted via an auxiliary rod, which is in turn screwed on the top of the scanner. As an imaging device, we used a DSLR of Canon, EOS 5D MarkII, which has been modified to enable NIR imaging. The RGB filter blocks all light below 830 nm, which perfectly satisfies our requirements. 

\subsection{Illumination variants}
\begin{figure}[ht]
\centering
\begin{minipage}{.5\textwidth}
  \centering
  \scalebox{.2}{\includegraphics{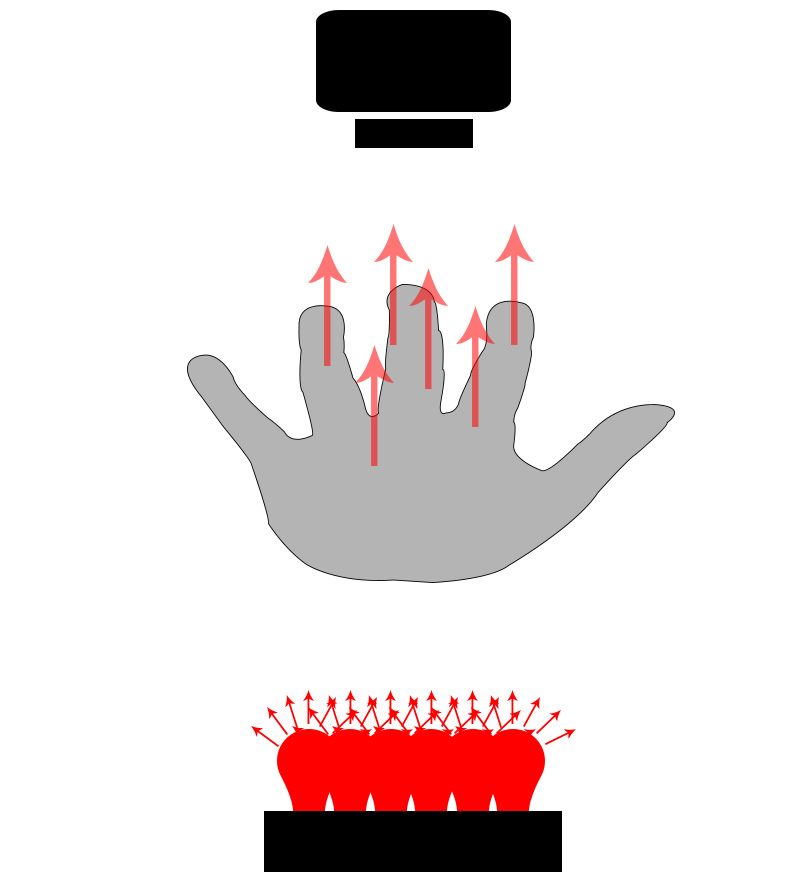}}
  \caption{Transmitted light}
  \label{fig:scanner_durchlicht}
\end{minipage}%
\begin{minipage}{.5\textwidth}
  \centering
  \scalebox{.2}{\includegraphics{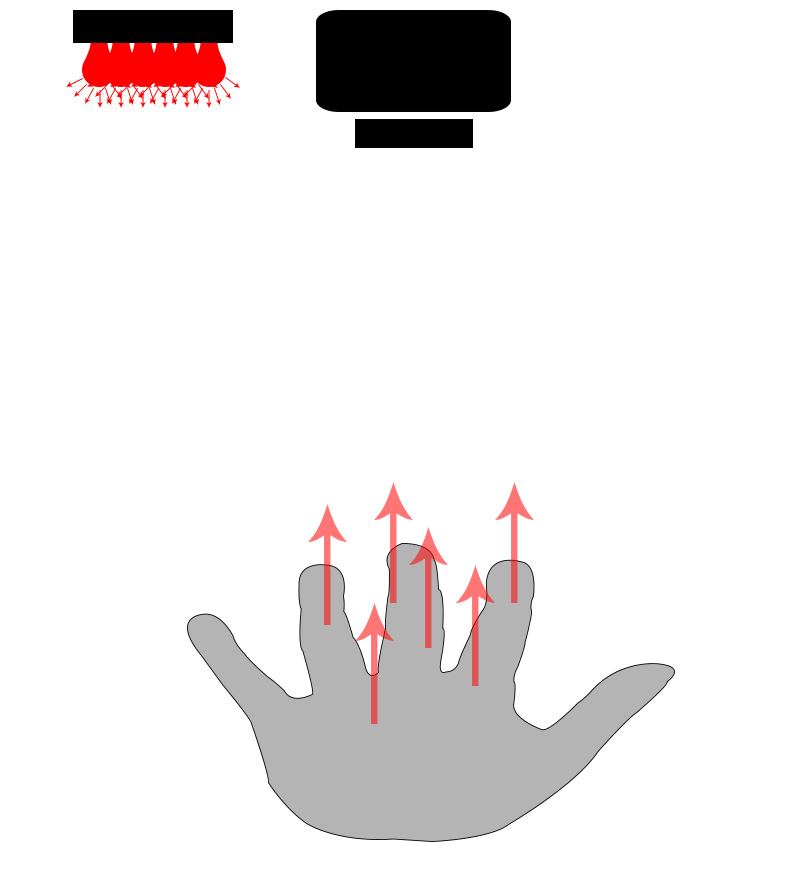}}
  \caption{Reflected light}
  \label{fig:scanner_auflicht}
\end{minipage}
\end{figure}

Probably the most distinguished property of our hand vein database is the illumination setup. Almost all established databases are rather concentrated on one illumination setup, instead of providing different ones. We tried to go one step further and decided to use two very different illumination variants, which can be seen in Figure \ref{fig:scanner_durchlicht} and \ref{fig:scanner_auflicht}. 

\newpage 

\subsubsection{Transmitted light}
This kind of illumination unveils the veins very clearly, the pattern can be seen without any processing. This is enabled by using a very concentrated NIR radiator right below the acrylic glass plate, where the hand is positioned during acquisition. The radiator consists of 50 NIR leds, which operate with a wavelength of around 950 nm. 

\begin{figure}[ht]
\begin{center}
\scalebox{.25}{\includegraphics{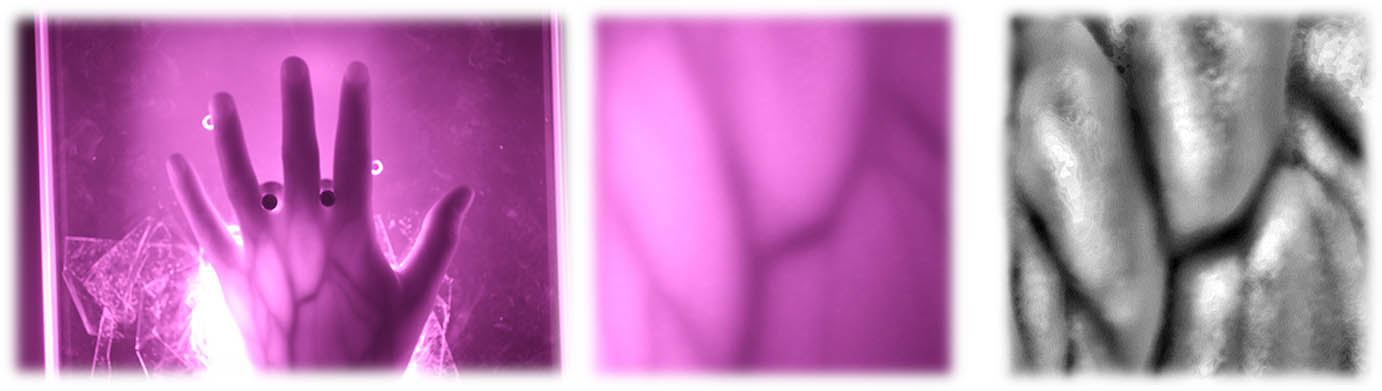}}
\caption{A subject's hand illuminated using transmitted light}
\label{fig:durchlicht_example}
\end{center}
\end{figure}

To illustrate the performance of this illumination variant, a sample from the database was chosen and is displayed in Figure \ref{fig:durchlicht_example}. Already the \textit{raw} captured image shows the veins rich in contrast. This contrast is underlined when looking at the ROI square (region of interest, center) and the preprocessed ROI square (right). 

\subsubsection{Reflected light}
The second illumination variant is very different to the first one. In comparison with transmitted light, we have a larger distance between hand and radiator, the latter is positioned at the same height as the camera. Further, we have a lower intensity of illumination, since only 6 NIR LEDs are being used. We tried two different NIR LED Arrays. Field tests have shown, that 940 nm LEDs perform better, so we decided to use this kind of LEDs. 

\begin{figure}[ht]
\begin{center}
\scalebox{.25}{\includegraphics{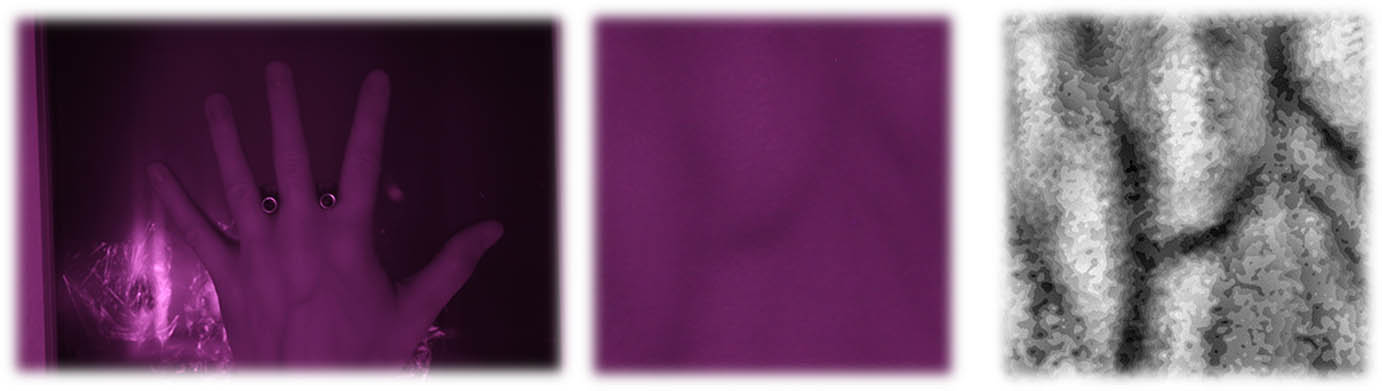}}
\caption{A subject's hand illuminated using reflected light}
\label{fig:auflicht_example}
\end{center}
\end{figure}

This suggests the assumption, that the hand will not be penetrated as much as in transmitted light case. Figure \ref{fig:auflicht_example} confirms this assumption. This illumination unveils only the veins, which are located right under the skin. Complex vein structures can only be reached when using a concentration as transmitted light enables.

\subsection{Acquisitions events and database details}
Since the project start in February 2014, we used several events, to collect data for our database. The first event we attended, was the \textit{Lange Nacht der Forschung 2014}, which took place on April 4th. Several institutes located in the state of Salzburg use this day to provide visitors the chance of seeing research live. As belonging to the \textit{Paris-Lodron University of Salzburg}, we had our booth at the \textit{Edmundsburg} in the city of Salzburg. During this evening, we recognized people's interest in the topic of hand veins. During almost all 5 hours of this event, we had a line of people, waiting for attending our live demonstration. In total, we were able to produce 43 sets of data.
\newline
We continued our database population on May 21st, at the open house event at the \textit{Faculty of Natural Sciences}, where we gained another 16 subject entries. The next event we used was the \textit{I-Day} on July 1st, at our \textit{Department of Computer Sciences}. This day is intended to inform visitors about the study of Computer Sciences at our department. Another 12 subjects visited our booth to participate in our live demo.  

\begin{figure}[ht]
\begin{center}
\scalebox{.25}{\includegraphics{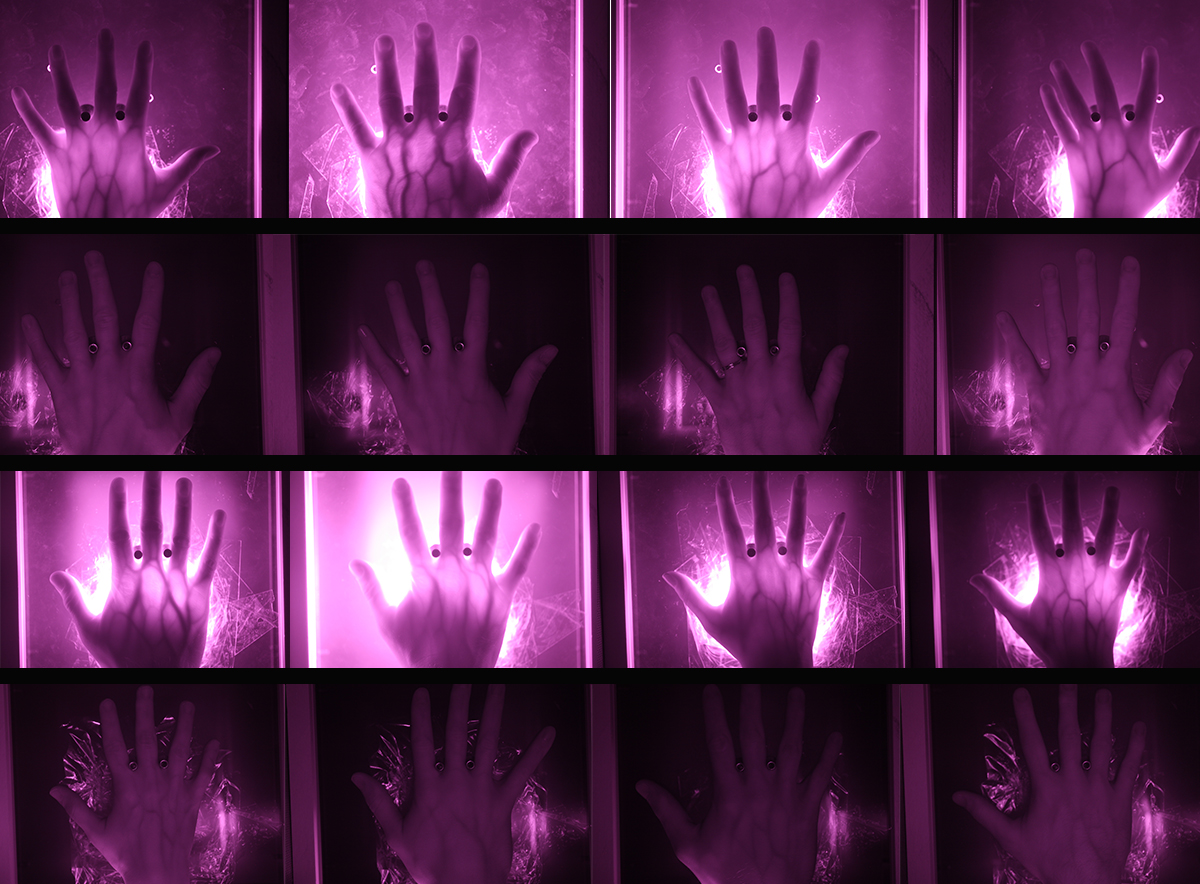}}
\caption{Sample images from our own hand vein database (arbitrarily chosen, row 1 - left hand transmitted light, row 2 - left hand reflected light, row 3 - right hand transmitted light, row 4 - right hand reflected hand)}
\label{fig:database_samples}
\end{center}
\end{figure}

Two final events took place on August 6th/7th, when we invited both students and faculty staff to become part of our database. Again, we experienced a very high interest and could increase the number of database entries by another 36 subjects. 
\newline
Summarized, we have a total number of 107 subjects in our hand vein database, which now consists of 1213 images, both in an uncompressed raw and compressed JPEG format. At the beginning, we always captured the left hand of subjects. But since the open house event at our department (since ID 72), we decided to record both hands, since we had more time available for acquisition. Each hand is captured three (or more) times, whereas the subject is told to do a pumping movement with the hand, before the third (or in general last) shot is taken. As additional meta information, we decided to record some subject-specific data: age, sex, weight and blood pressure. Fortunately, we were able to generate a database with a remarkable diversity in these meta characteristics.
To provide a quick insight, how these database images look like, see the extract in Figure \ref{fig:database_samples}.

The following table compares well-known hand vein databases with our own \textit{VeinPLUS} database mentioned rightmost. 
\newline

\hspace*{-1cm}
\begin{tabular} { l|l|l|l|l }
&CASIA \cite{casia-paper} & Bosphorus \cite{bosphorus-paper} & PolyU \cite{polyu-paper} & VeinPLUS (03-2015)\\ 
\hline
Images & 7200 &  1575 & 6000 & 1213\\
Subjects & 100 &  100 & 250 & 107\\
Contactless & yes &  yes & no & no\\
Illumination & front &  front & front & front and rear\\
Range[nm] & \begin{tabular}[l]{@{}l@{}}460,630,700,\\850,940\end{tabular}& - & 660,525,470,880 & \begin{tabular}[l]{@{}l@{}}940(r),~950(t)
\end{tabular}\\
Acquisition & front &  front & front & front\\
Regions & palm &  dorsa & palm & dorsa\\
Dimensions & 768*576 & 300*240 & 352*288 & 2784*1856\\
ROI & 128*128 & 100*100 & - & 500*500\\
\end{tabular}

\newpage
\section{Automated ROI extraction for hand vein database images}
Investigating samples of our hand vein database, it is obvious, which area forms our region of interest. Since we get the most vein information from the center of a subject's hand, we take a centered squared area that forms our ROI.

The process of ROI extraction has to be accomplished in an automated manner, that works the same way for all images. In the case of hand vein images, this means, that the extraction should always deliver the same squared area of the hand, no matter which variances are present in the image. Possible variances are rotation (the hand is present in a different angle), position (shifted in x or y direction), scale (the hand is closer or further away as seen from the sensor) and also exposure (over-/underexposure). So one might assume, that it takes a lot of work to find a suitable solution to the given acquisition environment. 

\begin{figure}[ht]
\centering
\begin{minipage}{.5\textwidth}
  \centering
  \scalebox{.3}{\includegraphics{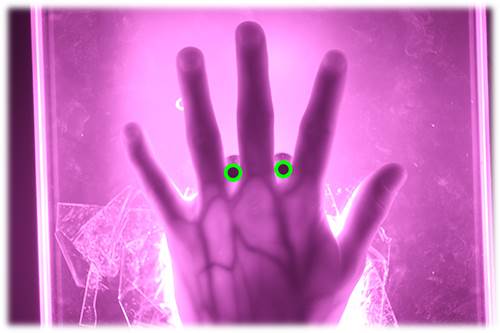}}
  \caption{Hough circles detected}
  \label{fig:roi_2}
\end{minipage}%
\begin{minipage}{.5\textwidth}
  \centering
  \scalebox{.3}{\includegraphics{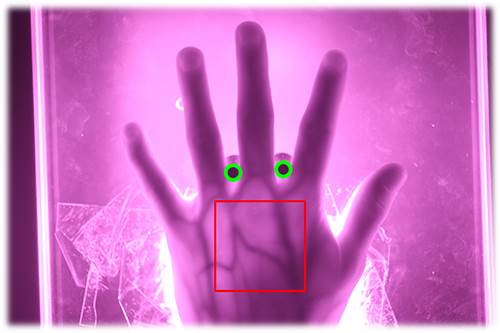}}
  \caption{ROI}
  \label{fig:roi_3}
\end{minipage}
\end{figure}

Fortunately, in the case of our hand vein database, there is one distinctive feature, which can be used to create the ROI area: the pegs.
The pegs were initially thought to act as a positioning assistant for subjects. But fortunately, these metallic helpers could be reused for ROI extraction. This is enabled by using the pegs' distinctive circular shape. As the pegs are the only circular shaped objects in the image, it is only needed to determine two circles in the image (see figure \ref{fig:roi_2}), which is accomplished by using \textit{Hough} transform algorithm. It's basic idea is to map edge information to a parametric curve model, to determine simple shapes like straight lines or circles. It's often used in combination with the established \textit{Canny} edge detector. The reason, why additional Hough transform is necessary, is that prior edge detection data contains incomplete shape structures. 

\begin{figure}[ht]
\begin{center}
\scalebox{.2}{\includegraphics{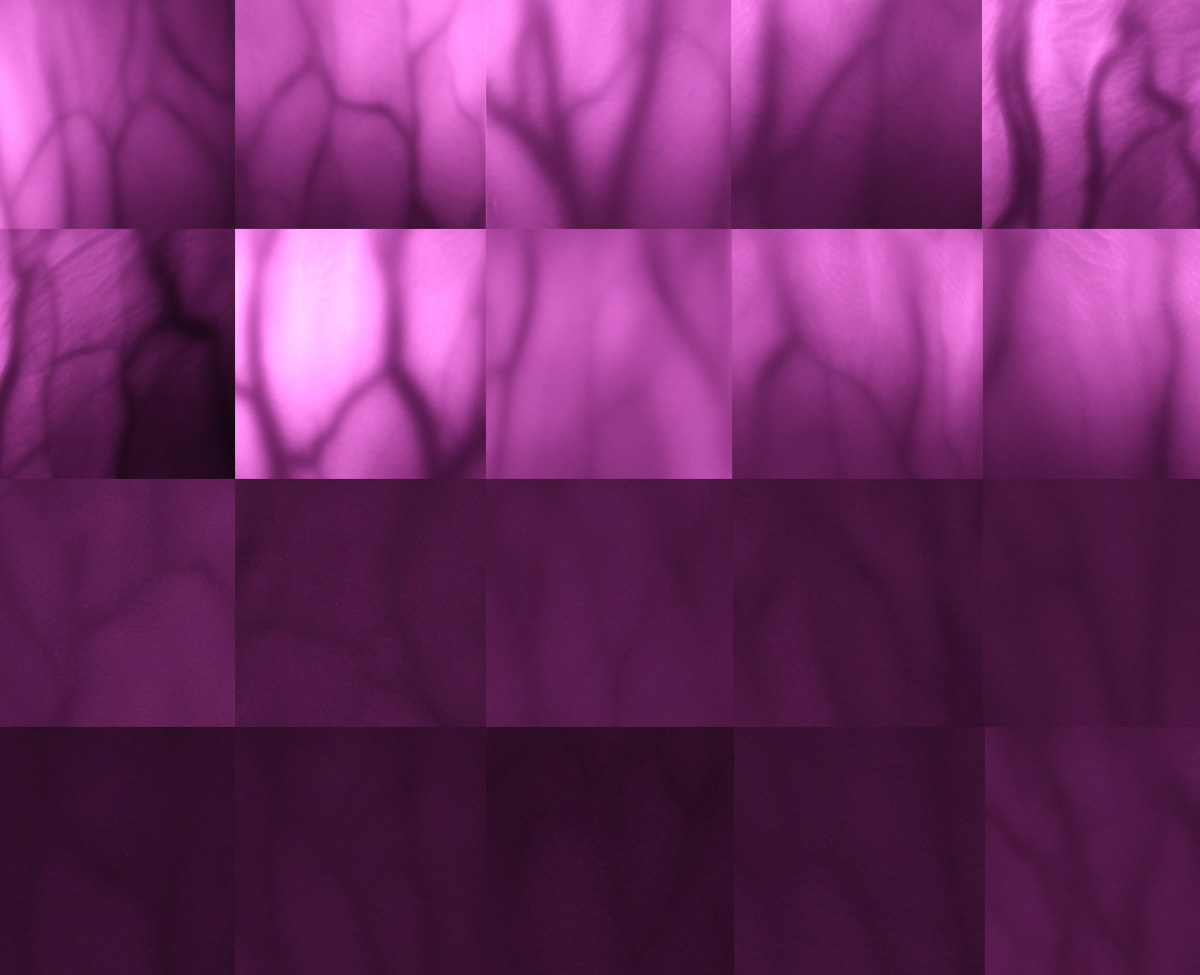}}
\caption{20 of 1213 ROIs in total (row 1,2 - transmitted light; row 3,4 - reflected light)}
\label{fig:roi_collection}
\end{center}
\end{figure}

From the circles' actual properties, we can use information to position the ROI square in the correct area. One of these properties is the circle's diameter. Since the pegs have a fixed size, we can make assumptions about the acquisition environment, when we detect different diameters. A greater diameter means, the subject's hand is closer to the sensor, e.g. when the acrylic glass plate is placed in a higher slot in the scanner box. In this case, we have to define a square with a bigger size, since the hand is also enlarged in scale (as observed by the sensor). The same observations hold conversely. 
\newline
The final result, the region of interest square, is shown in Figure \ref{fig:roi_3}. Since two different illumination variants were used, slight variations in the parameter choice were applied. But beside these variations, the same procedure could be reused. In general, the ROI extraction for a static environment is much easier to achieve, as opposed to a mobile environment. In Figure \ref{fig:roi_collection}, you see a collection of sample ROI images, produced by the extraction process explained in this section. 

\section{Conclusions}
Hand veins as a physiological feature in biometrical authentication systems are very promising. They show many benefits against existing well-established features, since they are counted among \textit{intrinsic biometrics}. Although they have been investigated only since around 2009, there are already several real world applications.
\newline
The acquisition takes place in the near infrared spectrum, since veins absorb near infrared light. Subsequent ROI extraction is applied, which produces an image of the most valuable vein area.  A department-hosted hand vein database was built, which contains images using both transmitted and reflected light illumination. Transmitted light illumination reveals the vein pattern very clearly without any processing. In our case, this was enabled by using a powerful NIR radiator, which is located right under the subject's hand. The second illumination variant, reflected light, is realized by using a radiator, which is positioned at top of the scanner, right next to the camera. The resulting images are by far not that rich in contrast as transmitted light enables and no deep veins are reached. 
As of March 2015, the database consists of 1213 raw and JPEG compressed original images. As part of the database we also provide ROI images, which were extracted by using the vein scanner's pegs as reference points. 

\section*{Acknowledgments}
This paper is an extract of my bachelor's thesis, so I would particularly like to thank Univ.-Prof. Dr. Andreas Uhl, who has supervised this project. I also want to thank Rudolf Schraml, a research fellow at WAVELAB, who was responsible for the design and construction of the vein scanner. If you have any further requests, don't hesitate to contact me via \textit{grusch.alex@gmail.com}.

\bibliography{refs}

\begin{thebibliography}{1}

\bibitem{casia-paper}
Hao et~al.
\newblock Multispectral palm image fusion for accurate contact-free palmprint
  recognition.
\newblock In {\em 15th IEEE International Conference on Image Processing, 2008.
  ICIP 2008.}, pages 281 -- 284. National Laboratory of Pattern Recognition,
  Institute of Automation, CAS, 2008.

\bibitem{bosphorus-paper}
Yuksel et~al.
\newblock Hand vein biometry based on geometry and appearance methods.
\newblock {\em Computer Vision, IET}, 5(6), 2011.

\bibitem{polyu-paper}
Zhang et~al.
\newblock An online system of multispectral palmprint verification.
\newblock {\em IEEE Transactions on Instrumentation and Measurement}, 59(2),
  2010.

\end{thebibliography}
\end{document}